\documentclass[nai]{iosart2x}

\usepackage{amsmath}
\usepackage{subfig}

\usepackage{graphicx}
\usepackage{amssymb}
\usepackage{epstopdf}
\usepackage{gensymb}
\usepackage{xcolor}
\usepackage{url}

\newcommand\og{\texttt{ochre\_gym}}
\newcommand\ogs{\texttt{ochre\_gym }}

\newcommand\sblines{\texttt{stablebaselines3 }}

\newcommand\pyt{\texttt{pytorch}}
\newcommand\pyts{\texttt{pytorch }}

\pubyear{2024}
\volume{0}
\firstpage{1}
\lastpage{1}

\begin{document}
\makeatletter
\let\put@numberlines@box\relax
\makeatother

\begin{frontmatter}

\title{Three Pathways to Neurosymbolic Reinforcement Learning with Interpretable Model and Policy Networks}
\runtitle{Differentiable interpretable neurosymbolic RL}


\begin{aug}
\author[A]{\inits{P.G.}\fnms{Peter} \snm{Graf}\ead[label=e1]{peter.graf@nrel.gov}%
\thanks{Corresponding author. \printead{e1}.}}
\author[A]{\inits{P.E.}\fnms{Patrick} \snm{Emami}\ead[label=e2]{patrick.emami@nrel.gov}}
\address[A]{Computational Science Ceneter, \orgname{National Renewable Energy Lab},
    CO, \cny{USA}\printead[presep={\\}]{e1}}
\end{aug}


\begin{abstract}
Neurosymbolic AI combines the interpretability, parsimony, and explicit reasoning of classical symbolic approaches  with the 
statistical learning of data-driven neural approaches.
Models and policies that are simultaneously differentiable and interpretable may be key enablers of this marriage.
This paper demonstrates three pathways to implementing such models and policies in a real-world reinforcement learning setting.
Specifically, we study a broad class of neural networks that build interpretable semantics directly into their
architecture. 
We reveal and highlight both the potential and the essential difficulties of combining logic, simulation, and learning.
One lesson is that learning benefits from continuity and differentiability, but classical logic is discrete and non-differentiable.
The relaxation to real-valued, differentiable representations presents a trade-off; the more learnable, the less interpretable.
Another lesson is that using logic in the context of a numerical simulation involves a non-trivial mapping from
raw (e.g., real-valued time series) simulation data to logical predicates.
Some open questions this note exposes include: 
What are the limits of rule-based controllers, and how learnable are they?
Do the differentiable interpretable approaches discussed here scale to large, complex, uncertain systems?
Can we truly achieve interpretability?
We highlight these and other themes across the three approaches.
\end{abstract}


\end{frontmatter}

\section{Introduction}

Approaches for developing increasingly general models based purely on data-driven statistical learning have shown great promise and achieved wide acclaim, most notably due to the recent success of large language models such as GPT-4. 
However, subsequent research applying close scrutiny to these approaches have revealed surprising limitations in their ability to perform basic logic, planning, and reasoning~\citep{mccoy2023embers,dziri2023faith,jelassi2023length,anil2022exploring,boix2023can,mitchell2023comparing}.
Neurosymbolic artificial intelligence (NSAI) is an alternative learning paradigm that attempts to integrate notions from classical symbolic AI with statistical approaches such as neural networks.
Proponents of NSAI argue that this combination may represent a fundamental leap in AI capability, which has helped attract attention to this approach (see reviews in \citep{garcez2023neurosymbolic,hitzler2022neuro,bouneffouf2022survey}). Recent workshops and summer schools~\citep{IBM23,Samsung22} capture the groundswell of interest in this field.
Some of the arguments in support of NSAI include: \textbf{a closer resemblance to human cognition}, as there is strong evidence that symbol processing is a core component of human mental processing \cite{marcus2003algebraic},  \textbf{interpretability}, as  explicit use of symbols within neural methods increases the transparency of how/why a model makes its predictions,  \textbf{factualness}, as encoding human knowledge and heuristics as helpful learning priors is enabled, and \textbf{reliability}, as correct reasoning can be guaranteed.

The goal of this note is to present three pathways to using differentiable and explicitly interpretable neurosymbolic policies in reinforcement learning and to illustrate such policies in simple
control tasks. 
Neurosymbolic reinforcement learning, or NSRL, aims to integrate neurosymbolic approaches within reinforcement learning (such as by replacing the policy or value network, or using differentiable planning to extract sequences of actions in model-based approaches).
In this work, we study NSRL for building energy management (BEM).
An example of a popular task in BEM is to control
building heating, ventilation, and cooling (HVAC). 
Buildings use a tremendous amount of energy  (in the range of 20-30\% worldwide~\citep{IEA23}), 
so strategies to reduce their energy usage, as well as strategies to intelligently integrate them into future power grids (e.g., thermal storage \citep{naderi2022demand,nelson2019residential}) are of great interest.
Furthermore, the current method of controlling them is through a series of logic-like rules (e.g., ``if temperature is greater than $20 \degree$ C, turn the
air conditioner on"), i.e., it is a symbolic, rule-based system, so NSRL is a natural fit for this domain.  
Our motivation behind grounding our discussion about the interpretability of neurosymbolic policy networks in a real-world problem is so that qualifiers such as ``more or less interpretable'' have meaning.

\subsection{Explicitly Interpretable Neural Network Architectures}

In this work, we wish to analyze different NSAI network architectures to better understand their promises and challenges for NSRL.
To that end, we pay particular attention to neural network architectures whose structure embodies rules and constraints we would like to apply. 
This approach has deep ties to cognitive science.  
In \cite{marcus2003algebraic}, Marcus
identifies three specific properties of cognition that AI systems should
support---an ability to: manipulate relations between variables,
form and utilize structured representations, and flexibly distinguish individuals from classes.
These days we realize each also needs to be \emph{learnable} (i.e., differentiable). How to support them in a differentiable framework is still an open question, 
but this line of attack suggests that this might
be achieved by designing neural network architectures that build in these properties, so that whatever is learned is guaranteed to have them.
The idea can also be summarized as designing architectures that have the ``concept of a concept" built-in (rather than \emph{specific} concepts built-in).
This is meant to force the construction of semantically relevant (e.g., interpretable) models.


These models can be directly implemented with standard deep learning frameworks (e.g., \pyt \cite{NEURIPS2019_9015}).
Thus formulated, these neurosymbolic models can be learned from data via standard stochastic gradient descent.  In principle, then, we have the best of both worlds, on the one hand a parsimonious, interpretable model, and on the other the differentiability that imbues neural networks 
with unprecedented power.  

\subsection{Illustrative test cases in BEM}

This work studies three pathways to integrate neural architectures that are explicitly interpretable into NSRL, in particular, as the policy that embodies the semantics we would like to employ.
 We study two particular models.  The first, Logical Neural Networks (LNNs) \cite{riegel2020logical}, were developed by IBM research.  The basic idea is to implement real-valued logic in a differentiable programming tool such as \pyt.
 The second approach we explore is differentiable decision trees (DDTs) \cite{silva2020optimization, tambwekar2023natural}.  A DDT can be thought of as a real-valued relaxation of a standard decision tree. 
 These two approaches are discussed in more detail in sections \ref{sec:lnn} and \ref{sec:ddt}.
 


We study these models as policies for NSRL using three demonstrations.
In the test cases, we will be working with simulations of buildings, not real buildings.  
Two of our examples utilize a 
state-of-the-art building simulator, OCHRE, developed at the US Department of Energy's National Renewable Energy Laboratory (NREL)
\cite{blonsky2021ochre,blonsky2021ochregithub}, along with our RL Gymnasium wrapper \og, that is specifically designed to explore problems with NSAI \cite{ochregym}.
The third will involve an extremely simple but fully differentiable simulation.

Our studies reveal and highlight both the potential and the essential difficulty of marrying logic and learning.
First, learning benefits from continuity and differentiability, but classical logic is discrete and nondifferentiable.  This essential tension causes numerical difficulties
and reduces interpretability.
Second, numerical simulations are not inherently described, thus controlled, via logical statements.  Bridging this gap
requires a form of ``translation"; ideally this bridge need not be hand-coded but is learned along with the logic-based controller.
Third, these approaches present scaling challenges; the differentiability makes them learnable, but considering ``all possible rules"
is still an exponentially large search space, which can rapidly become intractable even for differentiable methods.
These case studies and the lessons they offer will help guide our efforts as our understanding of how to build 
semantics into learnable architectures grows.

\section{Related Work}


There is a large body of work in NSRL (and NSAI in general), which we do not attempt to comprehensively review here. 
Illustrative examples of NSRL include hybrid approaches to perception and planning such as \cite{garnelo2016towards}, where neural networks learn to convert images to classical planning problems, and programmatic RL \cite{verma2018programmatically}, which solves control problems by allowing controllers to be learnable programs.

We briefly review previous work using both DDTs and LNNs in the context of reinforcement learning.
DDTs were introduced in \cite{suarez1999globally}, where they were used for classification and regression tasks.  They were
first applied to reinforcement learning in \cite{silva2021encoding}.  This paper also introduced the ``warm start" concept that allows
initializing a DDT in an approximately optimal state.  They explored several domains from the standard CartPole to a custom
wildfire tracking application.  This approach was extended to allow for natural language specification of the warm start policy in \cite{tambwekar2023natural}.
Other relevant applications include \cite{kalra2022interpretable}, where the authors use DDTs to learn reward functions from human feedback,
\cite{ding2020cdt}, in which DDTs have been used to explain deep RL policies, and \cite{basaklar2023dtrl}, where DDTs are used for 
task scheduling on microchips.

The notion of relating logic to neural networks goes quite far back (e.g., \cite{ludermir1991relating,kijsirikul2005first}). LNNs in the form 
we use in this paper were introduced by IBM in \cite{riegel2020logical}.  Since then, they have been extended and used for several studies. 
In \cite{sen2022neuro} LNNs were extended to first-order logic and applied to a variety of knowledge base completion benchmarks.  A first application
to reinforcement learning is \cite{kimura2021neuro}, in the context of a simple text-based game.
In \cite{agravante2022learning} LNNs are used to build a world model of a text-based game, from which a classical planning problem can be 
built and solved to choose the ``logically optimal action".  This approach is the inspiration for the model-based RL scheme described in 
section \ref{sec:mbrl}.  


The main distinguishing features of the present work are: application of LNNs and DDTs to building energy management rather than toy problems; 
integration of a differentiable interpretable approach
directly into a standard RL framework; exploration across three different RL paradigms;
recognition that these formulations are instances of the much broader concept of building semantic structure into learnable architectures.

\section{Background}
In this section, we first introduce differentiable decision trees (Section~\ref{sec:ddt}) and logical neural networks (Section~\ref{sec:lnn}). Then, we review the various NSRL paradigms considered in our test cases (Section~\ref{sec:nsrl}). 

\subsection{Differentiable Decision Trees}
\label{sec:ddt}
Standard 
decision trees for, say, a classification task, are generally built in a greedy top down manner in which the most ``important" attribute (typically from an information-gain perspective)
is chosen first, along with a threshold value, which splits input data according to the values of this particular attribute.  The process is repeated recursively to some depth,
resulting in a tree of decision nodes, each of which compares some attribute to some threshold.  At the bottom of the tree, the leaves contain the value of the predicted class for an
input vector whose path through the tree leads to that leaf.
In a DDT, instead, the decision nodes compare a \emph{linear combination of all the attributes} to a threshold, and the leaves give probabilities over all possible classes.  To make the 
decision nodes differentiable, the threshold is replaced by a sigmoid function of the difference between the linear combination and the threshold value.  In a DDT, the weights of the linear
combination, the threshold values, the strength of the sigmoid (how close to a step function it is), and the leaf probabilities are all learnable parameters.  

\begin{figure}%
    \centering
    \subfloat[\centering Nodes and levels of a 4-leaf tree labeled by the binary multi-index according to depth and branching used in the text.]
    {{\includegraphics[width=7cm]{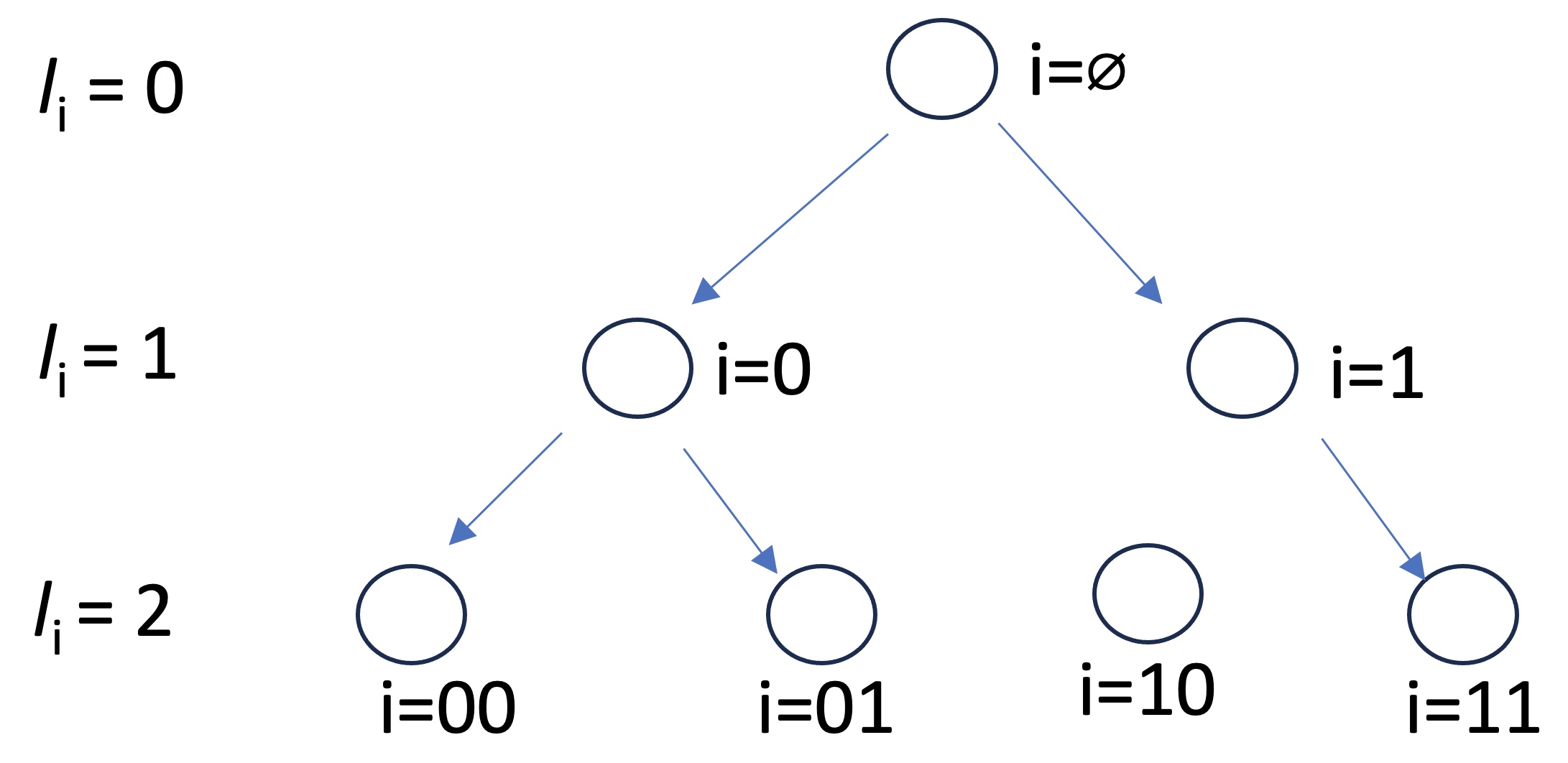} }}%
    \qquad
    \subfloat[\centering ``Warmstart" DDT corresponding to our tuned rule-based `pre-cooling' controller.  True conditions result in taking the left branch.  Note the redundant left branches, retained for implementation reasons. Here, $P_{cur}$ = current price of power, $P_{fut}$ = future price of power, $T_{set} $ = HVAC temperature setpoint.]
    {{\includegraphics[width=7cm]{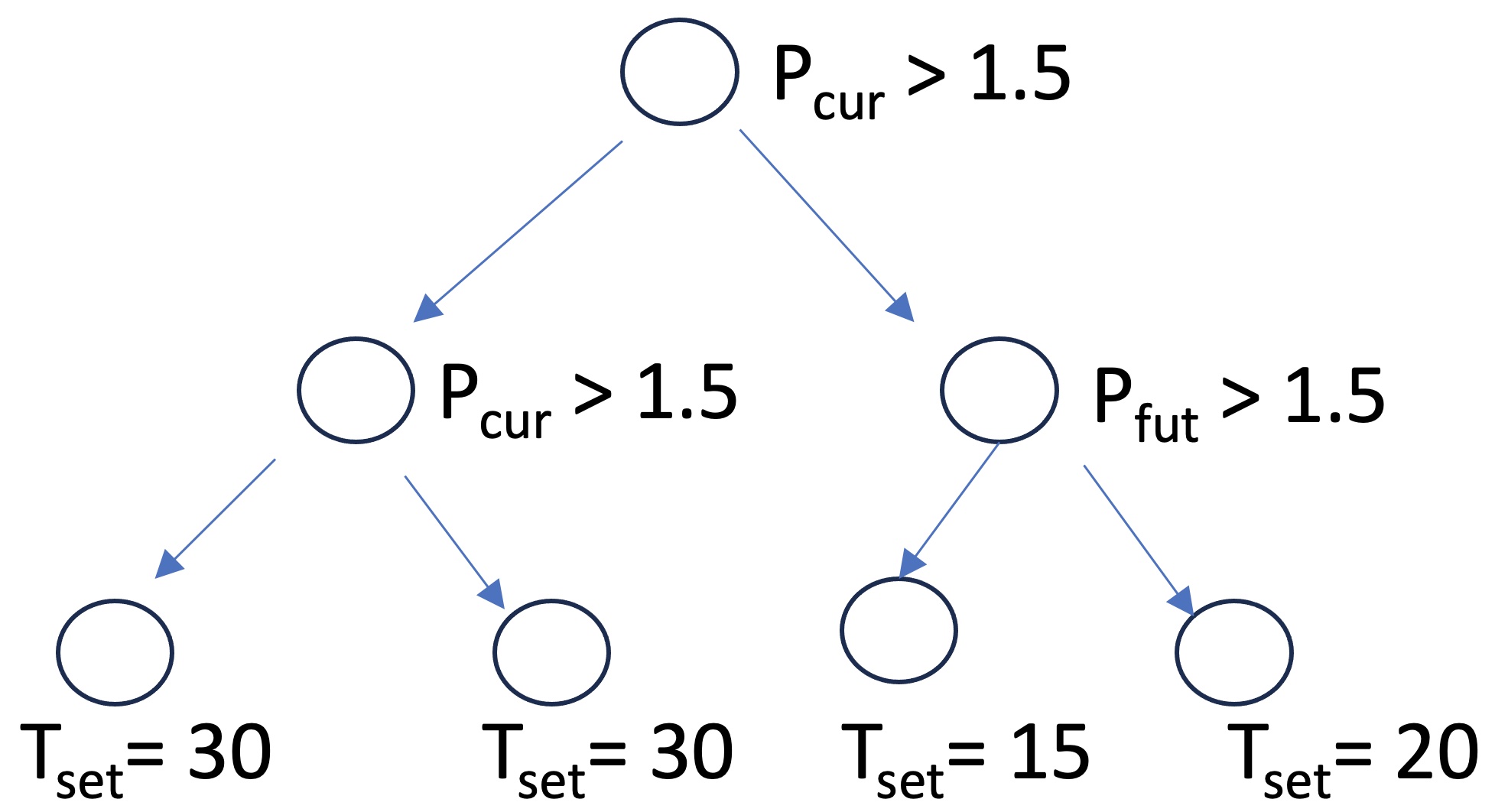} }}%
    \caption{Details of a 4-leaf DDT.  (left) the mathematical formulation's indexing scheme  (right) Our baseline RBC controller (see section \ref{sec:ddtsetup}) as a DDT from which we can initialize learning.}%
    \label{fig:ddtdetails}%
\end{figure}

Formally, a DDT can be described as follows (roughly following \cite{silva2020optimization,silva2021encoding}).
We identify nodes in the tree via a multi-index $i \in \{0,1\}^{l_i}$, where $l_i$ is the depth of the current node.  For example, the
nodes of a 4-leaf  tree can be labeled such as shown in Figure \ref{fig:ddtdetails}.  Now, given input $x$ we define the probability of arriving
at node $i$ recursively via
\begin{equation}
    p_i(x) = \sigma(\gamma_i (w_i \cdot x - c_i)) * p_{i_{par}}(x), \quad \quad p_{\varnothing} = 1,
\end{equation}
where $\sigma$ is the sigmoid function, $\gamma_i$ is the strength of the sigmoid, $w_i$ is the vector of weights for this node,
$c_i$ is the threshold (``comparator") value, and $i_{par}$ is the index of the parent node of $i$.  The probabilities
cascade recursively to the leaf nodes.  Meanwhile, the leaf nodes contain a vector of probabilities of each of a 
 set $\{a_k\}$, $k = 1..N_a$ of $N_a$ output values for that leaf ($N_a$ is the number of 
discrete actions from which we will build a final continuous action value
required by the soft actor critic (SAC) RL algorithm).
At each leaf node ($l_i = 2$ in the figure) let the vector of
action probabilities be
$y_i = \{q_i^k\}$ for $k = 1..N_a$.
Considering all paths through the tree and their probabilities combined with the action probabilities, the overall probability of each action level is
$r = y_i \cdot p_i \in \mathrm{R}^{N_a}$.
To get a continuous action value, finally, we compute 
the  ``soft-action"  via
$a(x) = \sum_{k=1}^{N_a} r_k a_k$.
This particular relaxation is applicable when the discrete action values represent quantized levels (in order of increasing magnitude) of a continuous interval. 

\subsection{Logical Neural Networks}
\label{sec:lnn}
The foundation
of LNNs is real-valued logic, of which there are many variants (see, e.g., \cite{riegel2020logical}, Appendix B).  Real-value logic allows us to formulate the ``truth-table" of standard logical connectives (e.g., AND, OR) as functions on the unit square with range [0,1].  We define a real value truth threshold, $\alpha \in [0,1]$. A statement
is true if its value is greater than or equal to $\alpha$, false if less than or equal to $1-\alpha$, and unknown otherwise. The power of LNNs comes from allowing for these truth-tables to be \emph{learned} rather than fixed.  Thus, one can formulate hypothetical logical 
rule-templates and learn their exact semantics from data.  For example, a model for the day's weather might begin as ``cold AND rainy", but given examples of true facts about
the day (``cold and rainy", ``cold and dry"), an LNN can learn that the true model of the day's weather is simply ``cold". The AND-gate learns to ignore its second 
argument, ``rainy", as it does not help describe the true state of affairs.  Importantly, this learning process happens through gradient descent of a loss function
exactly analogous to (implemented in the same tools as) standard NNs.

The mathematics of LNNs is slightly more complicated than DDTs (see \cite{riegel2020logical} (and \cite{graylnnweb} for a gentler introduction)).  
Logical gates in LNNs are similar to decision nodes in DDTs. Both can be described by applying a ``threshold-like" function to a weighted sum of attributes.  I.e., we write the output truth value $y$ of a gate as $y = f(w \cdot x - \theta)$, for
the nonlinear activation function $f:\mathbb{R} \rightarrow [0,1]$.  Variants of real-valued logic are largely implemented through choices of $f$.  The sigmoid, $f = \sigma$, as in DDTs, is a common choice.  The character of the gate is given by constraints on the weights $w$ and bias $\theta$.  For example, 
AND-like and OR-like semantics are given by the constraints listed in Table \ref{tab:lnncons}.

\begin{table}
    \centering
    \begin{tabular}{|c|c|c|}
    \hline
      Gate   & Semantics  &  Constraints\\  \hline
      AND   & ``One input False means output False" & $\forall_i, \sum_j w_j - w_i \alpha  - \theta \leq f^{-1}(1-\alpha)$  \\
         &   ``All inputs True means output True" & $\sum_j w_j \alpha   - \theta \geq f^{-1}(\alpha)$ \\  \hline
      OR   & ``One input True means output True" & $\forall_j, w_j \alpha   - \theta \geq f^{-1}(\alpha)$ \\
         & ``All inputs False means output False" & $\sum_j (1-w_j) \alpha  - \theta \leq f^{-1}(1-\alpha)$ \\  \hline
    \end{tabular}
    \caption{LNN AND- and OR- like semantics and corresponding linear constraints on gate weights, $w$, and bias, $\theta$.
    The top formula, for example, is derived by computing $f(w \cdot x - \theta)$ assuming the $i$th element of $x$ is False and the others are True,
    which results in $f(\sum_j w_j - w_i \alpha  - \theta)$, which must output a False value, i.e., one that is $\leq 1-\alpha$.  Applying $f^{-1}$ yields
    the constraint.
    }
    \label{tab:lnncons}
\end{table}

More complicated logical formulas are built by composing the primitive AND, OR, and NOT gates.  We can think of this, similar to the DDT case, as recursively feeding the output of one operator to the input of another.  When we write logical formulas explicitly, we select the features of interest ``by hand".  This amounts to preselecting the weights of the operations.  However, LNNs can learn the weights from data. 
A popular strategy is to start with equal weights over all the features and allow the LNN machinery to learn which to ignore, i.e. which weights to set to zero.
The combination of initially including all the attributes and then chaining the operators makes LNNs look very much like DDTs, i.e. a cascade of thresholded 
weighted comparisons.  These formulas, overly general because they include irrelevant attributes, become rule ``templates" from which we can start the 
learning process.

\subsection{RL Paradigms}
\label{sec:nsrl}
For purposes of this paper we divide approaches to reinforcement learning (RL) into three classes: model-free RL (MFRL), model-based RL (MBRL), and an approach
relying on differentiating \emph{through} a simulator that has become known as differentiable predictive control (DPC) \cite{drgovna2022differentiable}.  
Figure (\ref{fig:threemethods}) sketches pseudocode of generic versions of these three modalities, summarized as follows:
\begin{figure}[h]
\includegraphics[width=6in]{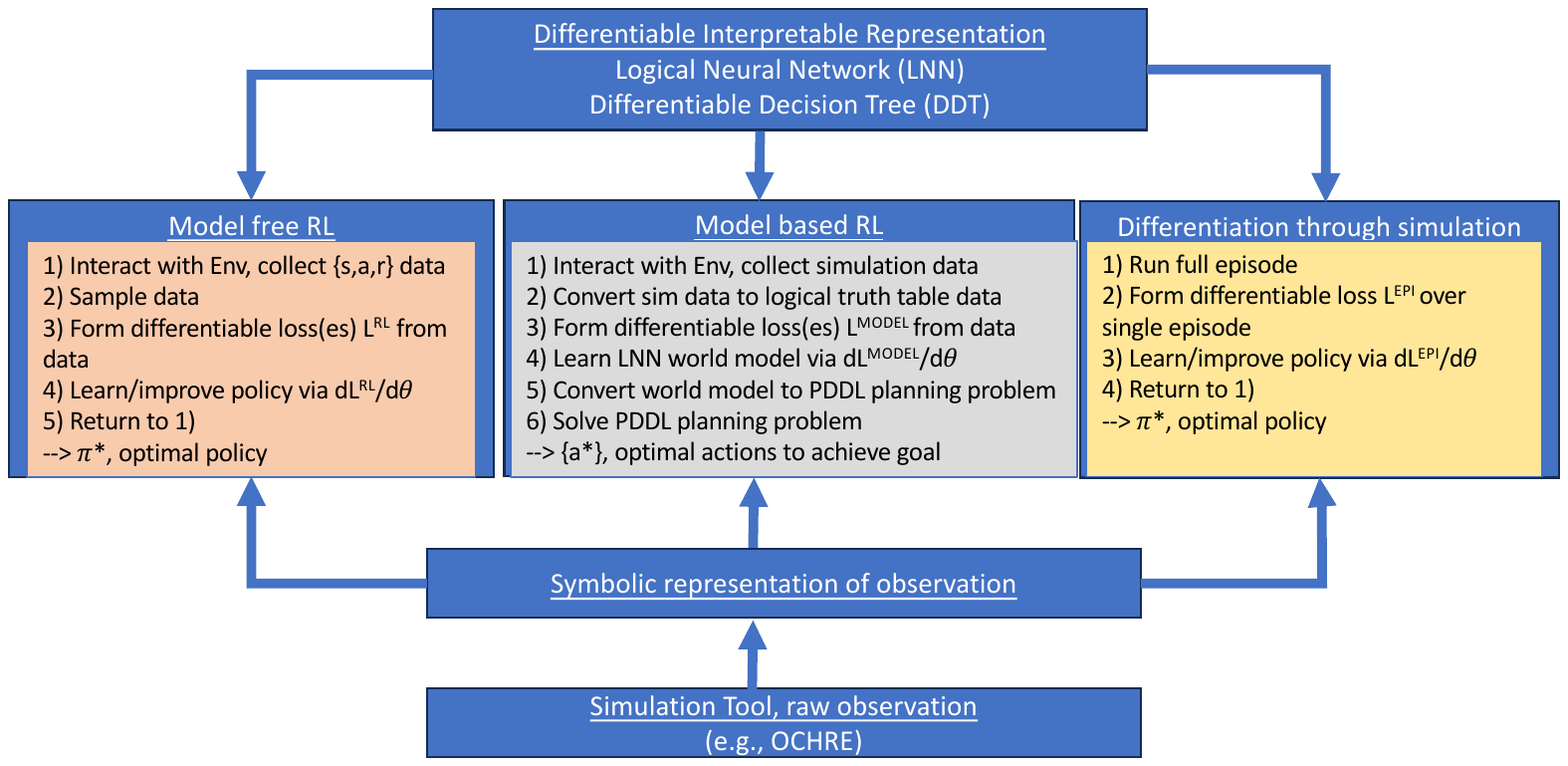}
\caption{Pseudocode of three methods to achieve symbolic RL explored in this paper. 
(left) Model-free RL; in our case the policy is a DDT (middle) Model-based RL with LNN 
model used to build and solve classical planning problem. (right) Direct optimization  integrating 
a differentiable simulation and LNN policy.}
\label{fig:threemethods}
\end{figure}

\textbf{Model-free Reinforcement Learning:} In model-free reinforcement learning (MFRL), the control policy is directly learned by interacting with the system.  This is the most common form, easily implemented and studied in a wide variety of contexts \cite{degris2012model, ccalicsir2019model}, including
building energy management \cite{nagy2023ten,dolatabadi2022novel, fu2022applications, hou2023model, zhang2022building}.
In this paper we will integrate  DDTs with the \sblines (SB3) RL framework and use it to solve control problems with \ogs using the soft actor critic (SAC) \cite{haarnoja2018soft} method.

\textbf{Model-based Reinforcement Learning:} In MBRL, the agent learns a  world model by interacting with the system; the model is then optimized to control the system.  In most cases it is assumed that the world model is a neural network.  See review in \cite{luo2022survey}.  It becomes a subsequent body of 
research how to use the model to optimally control a system \citep{liu2020constrained,pineda2021mbrl,chua2018deep}.  
In our case, we will demonstrate a pipeline 
that involves learning a symbolic world model in the form of the pre- and post- conditions of a classical planning problem (e.g., written in PDDL).  
This problem can then be directly and exactly solved by a classical planner.
Classical planning often makes use of a STRIPs~\citep{fikes1993strips,asai2020learning}
 or similar formulation of a planning problem that consists of an initial state, a goal
state, and a set of possible actions.  In STRIPs, each action is specified according to its pre- and post- conditions.  Preconditions are divided into
statements that must be true (``positive preconditions") for the action to be executed and statements that must be false (``negative preconditions").  Similarly,
postconditions consist of statements that are made newly true by the action and statements that are made newly false.  
We use this STRIPS-based planning problem description as the symbolic model of the system we wish to control.

\textbf{Differentiable Predictive Control:} Our third approach, sometimes called differentiable predictive control (DPC) \citep{drgovna2022differentiable,drgovna2022learning} relies on a simulation that itself is differentiable (e.g., implemented in \pyt).  If the control scheme is also differentiable, such as the case here with DDTs and LNNs, then we 
can integrate the controls problem and the dynamics and solve for the optimal control policy all at 
once. Here we will describe a simple differentiable temperature regulation
simulation implemented in \pyts and demonstrate the learning of an optimal LNN-based control policy.

\section{Experiment}

In this section, we study model-free RL approaches with DDTs. Specifically, we present the results of an HVAC control experiment where the performance of 1) rule-based control, 2) model-free RL with a DDT policy, and 3) model-free RL without a DDT policy, are compared. 

\subsection{OCHRE Gym}
\label{sec:ochre-gym}
Here, we introduce the details of the RL environment used for this experiment.
This experiment uses the RL environment \ogs\cite{ochregym}, which is an abstraction implemented on top of OCHRE~\citep{blonsky2021ochre}, a Python-based residential building energy simulator, for RL research. 
OCHRE uses a 5-zone building representation constructed from multiple resistance-capacitance  models to capture how a basic residential building consumes energy. 
It includes physics-based models for household devices including HVAC equipment, water heaters, and more. 
OCHRE has been designed to emulate the high-fidelity EnergyPlus~\citep{ePlus} building simulator as a reduced order model,
making it a useful and flexible EnergyPlus surrogate for research.

Our platform \ogs provides an OpenAI Gym API for OCHRE and
 is designed to be flexible; that is, the observation space, action space, reward functions, and underlying environment (the EnergyPlus building configuration) are all customizable. 
The default \ogs observation space is designed to have the same diagnostic variables that OCHRE returns after each BEM control time step (the user is able to remove unwanted variables, however).
There are around 50 such variables; prominent examples include the current outdoor temperature, the current indoor temperature, the current total electric power consumption, the current HVAC heating electric power consumption, and the current energy price. 
Users of \ogs can modify this to ignore unwanted variables to simplify the environment, if desired.
Agents act by controlling the various types of equipment in the structure, e.g., an HVAC heater, an HVAC cooler, a water heater, etc.
Each equipment can be controlled by setting the setpoint or the duty cycle (which are continuous actions), and can be ``turned off'' with a discrete binary ``load fraction'' action.
The \ogs reward function supports three common demand response scenarios (real-time pricing, time-of-use pricing, and power-constrained operation).  In each case, the objective is to minimize the total energy costs over the control period (the episode) while maintaining the temperature inside a comfortable range.

\subsection{Setup}
\label{sec:ddtsetup}
\textbf{BEM problem description: } This test case examines the ability of an agent to learn to control HVAC  under the ``time-of-use" (TOU) pricing reward scenario in \og, where
the unit price of power is 10 times higher between noon and 6 pm.
Agents train on the month of June and are tested on all months,
where the conditions (e.g., weather) differ.  

\textbf{Implementation: }We adapt the DDT model 
from \cite{silva2020optimization} to be 
the actor network within \texttt{stablebaseline3}'s \citep{raffin2019stable} Soft Actor Critic (SAC) implementation.   We chose SAC because we found it worked better than 
other deep RL algorithms on this TOU scenario.  Also, the analysis in \cite{silva2020optimization} suggests that DDTs work better as the
policy than the Q-function.   
Several technical hurdles had to be addressed to instead make use of a DDT.
The \texttt{SACActor} class was subclassed to replace the default network with our DDT.
The \texttt{SACPolicy} class was subclassed so the \texttt{make\_actor()} function could make our custom, DDT-based Actor.  
The \texttt{SAC} class was subclassed to include our custom policy in the available policies.
The SAC algorithm requires continuous actions, whereas natively DTTs produce discrete actions (each possible leaf node value is
a distinct action).  To address this, as described above we used ``soft actions" in which the action value is a weighted combination of a set of discrete action values.

\textbf{Baselines: }We also trained a deep RL model with the same SAC method, as well as coded a simple rule-based
controller (RBC).  The RBC adopts the strategy of ``precooling"; the HVAC setpoint is set especially low for several hours before the the TOU period, and
the setpoint is set especially high (essentially turning the HVAC off) during the TOU period.  This strategy is widely studied and
easily implemented as rule-based control \cite{naderi2022demand}.  More specifically, our rule based controller is
\begin{align*}
&\texttt{If EnergyPrice} > 1.5 & \texttt{then HVAC Setpoint } = 30\degree C\\
&\texttt{Elseif FutureEnergyPrice} > 1.5 & \texttt{then HVAC Setpoint } = 15\degree C\\
&\texttt{Else} &  \texttt{HVAC Setpoint } = 20 \degree C\\
\end{align*}
The specific values of 15 and 20 were tuned to optimize the controller over June weather data.  The value of 30 is effectively turning the HVAC off during 
the time when energy prices are high.  

The input variables (observation space) upon which the controllers make their decisions vary between controllers.   Table \ref{tab:nsrlvars} summarizes 
the variables each controller has access to. The RBC controller only uses the current energy price 
and future average energy price, from which a rule such as ``if the current price is low but the future price is high, then precool" can be implemented.
The DRL control has access to much more information (much of it probably irrelevant), in total 26 variables, several of which are the exact energy
price for a ``lookahead" horizon at various time points.  The DDT controller has access to current and future prices, like the RBC, but also the indoor and 
outdoor temperature at any given time.  

One of the potential benefits of a logical approach is that we can incorporate knowledge.  Here, that constitutes knowledge
about the quite effective RBC solution.  We can, in fact, ``warmstart" our DDT with the RBC exactly,  
as shown in Figure \ref{fig:ddtdetails}.

\begin{table}
    \centering
    \begin{tabular}{|c|c|c|}
    \hline
      Method   & Obs Space Size  &  Obs Space Attributes\\  \hline
      DRL   & 26 & All OCHRE attributes  \\
      DDT   & 4 & Indoor temp, Outdoor temp, Current energy price, Future energy price  \\
      RBC   & 2 & Current energy price, Future energy price  \\ \hline
    \end{tabular}
    \caption{The number and nature of the variables supplied to each of the controllers compared in our model free RL experiments.}
    \label{tab:nsrlvars}
\end{table}

\subsection{Results} 
\label{sec:ddtresults}
\begin{figure}[h]
\includegraphics[width=6in]{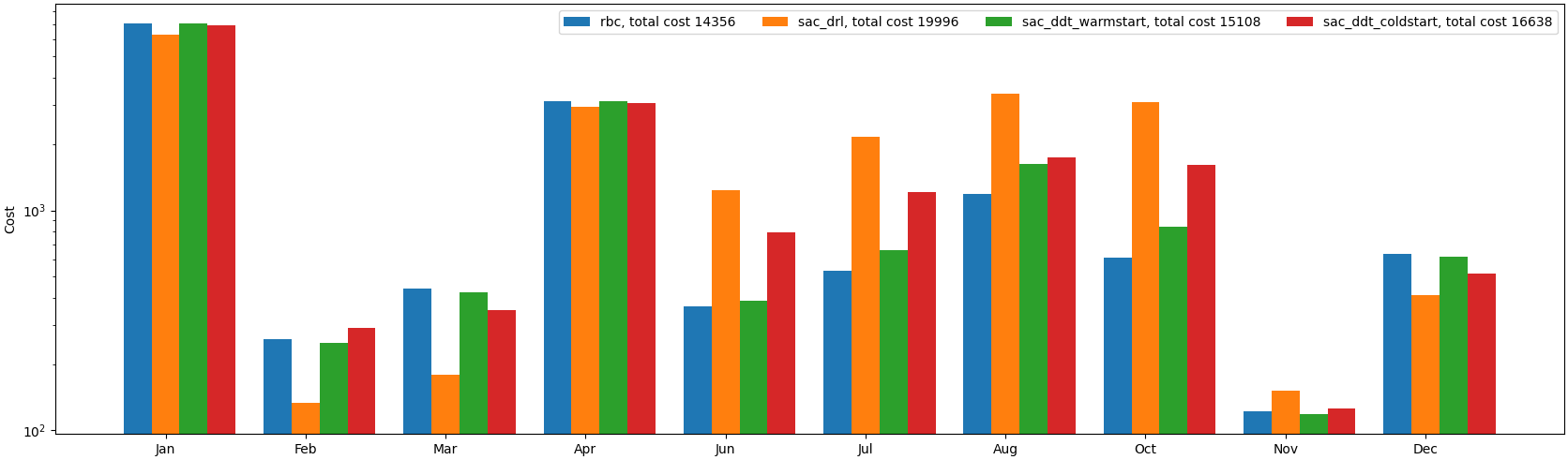}
\caption{Comparing rule-based, deep RL, and DDT based controllers for a 30 day TOU task.  The DRL and DDT controllers were trained 
(and the RBC controller tuned) on data for June.  The figure then shows the 30 day cost when these controllers were used, unmodified, across
9 other months.  The ``warmstart" DDTs are (approximately) initialized with the RBC ``precool" policy, whereas the ``coldstart" DDTs 
are randomly initialized.}
\label{fig:ddtresults}
\end{figure}

A summary of results is shown in Fig. \ref{fig:ddtresults}. Some observations regarding these results follow.
Overall, the RBC outperforms the DDT and DRL controllers.
The main problem we encountered is that the DDT model, which includes a cascade of sigmoid functions, in particular,
presents a difficult challenge for the standard stochastic gradient descent employed by SB3.  
SGD benefits from an underlying smoothness in the exact loss function, so approximating it by samples is still
effective.  But when our model, in this case the DDT, is continuous but ``wants to be discrete", which is evinced 
here by the ubiquitous use of sigmoids to approximate step functions, we are on dangerous ground.  Sampling and
approximate discontinuity are a challenging combination.

Nonetheless, we can still make some positive points about SAC with DDTs.  The rule based controller is not adaptive, whereas SAC with DDTs is.  We see from the 
data that the RBC performs badly in cool months.  This is likely because it is turning on the HVAC when it is manifestly not necessary.
As humans we can easily solve this by adding more rules (``don't precool if it's cold outside").  But we note that as situations get more complex,
this approach does not scale.  The whole idea is that we should be learning rules from data that do not require human intervention.
Finally, we note, deep RL (without DDTs) is easy to implement.  But the results here suggest that it is otherwise the least desirable approach,
as it displays no better performance than the DDT policy, without the benefit of any sort of interpretability (and subsequent flexibility for,
e.g., composability) that symbolic policies offer.

\textbf{Intepretability:} Unfortunately, interpretability and differentiability fight against each other in this framework.  Interpretability needs sharp decision boundaries, not a cascade of linear combinations minus thresholds fed through sigmoids.  But 
 learnability, reliant on differentiability, requires  smooth functions throughout.
  Another problem is that we had to generate continuous action values from an inherently discrete model (DDT).
 The use of the soft-action to do so has likely exacerbated the difficulty.
 The `discretization' of learned DDTs has been discussed in \cite{silva2020optimization}, where weights are `snapped' to the nearest integer (0 or 1),
 with excellent results for Cart-pole and other test cases.  In our case, our `warm-start' solutions general stay close enough to the
 rule-based starting point to be amenable to this approach.  The non-warm-started solutions do not.  We have experimented with several 
 regularization strategies that add a term to the SAC loss function to encourage integer solutions, with mixed success.  Such a term can look like
  $  |w|_p + (|w|_1 - 1)$,
 where $w$ are attribute weights and we use the $p$-norm (for $p>>1$) to push all the weight toward one attribute and the $1$-norm to force this
 single weight to be 1.  
 In math programming, e.g., the branch-and-bound method for integer programming, the use of relaxations is standard. There, though, the relaxed solution
 is compared to feasible (i.e., integer) solutions within an algorithm that is guaranteed to converge to an optimal and feasible solution.  
 To the extent DDTs can be considered relaxations, they are uncontrolled, that is, never constrained to be integer; they thus invite various
 ad hoc truncation and regularization approaches.
 
The results in Figure \ref{fig:ddtresults} 
show that SAC with DDT and the ``warm start'' initialization achieves a training month (June) cost that is roughly equal to the June RBC cost.  However, we found 
that this strategy does not \emph{always} work,  probably due to issues of statistical nonconvergence (not shown). 
 Especially for large learning rates or RL settings encouraging exploration, the solver immediately makes a big step 
 away from the RBC rule, at which point it is no better than an uninitialized solution.
 DDTs are not unique with respect to this phenomenon; we have seen a similar issue occur in methods for discovering a protein with a specific function by ``warm starting'' from a known ``good'' protein~\citep{emami2023plug}.  
 Methods that fail to properly constrain the optimization process to stay ``close'' to the starting protein tend to quickly end up searching within regions of the vast space of protein sequences populated by unnatural, nonfunctional proteins.
This implies more sophisticated techniques may be needed to get ``warm starting'' working reliably for DDTs for building control.

\section{Test Case 1: Model-based RL with LNNs and classical planning}
\label{sec:mbrl}

\subsection{Setup}

Here, we demonstrate an 
approach based on inductive logic programming (ILP)~\citep{muggleton1994inductive}, in which a LNN world model is learned in the form of 
a classical planning problem~\citep{agravante2022learning}.
For each possible action, we will learn logical expressions for its pre- and post- conditions.  These logical statements can be expressed directly in the PDDL planning language.  Then the goal state and initial state can be added and the problem solved exactly by a classical planner.
The workflow of our example of this process using  OCHRE is shown in Figure \ref{fig:loa-demo}.  For this problem, imagine you are in a house, and there is a 
mystery switch; you don't know what it does.  To figure it out, you can probe the effect of this switch in various settings and derive conditions under
which it has an effect, and what that effect is.  These correspond to the pre- and post- conditions necessary to implement a STRIPs style plan in PDDL.
Our toy setting is extremely simple and is designed simply to demonstrate that this chain of modeling is possible and that the learnability of LNNs can
be leveraged to enable it.   

To make use of LNNs, we need to define a ``vocabulary" such that simulation values can be mapped to
logical predicates (see Figure \ref{fig:loa-vocab}).
Our example considers two logical predicates, ``cold", and ``red" that are to be characteristics of a day (``red" is, by design, a spurious input; it will never change the state).  The action in question is \texttt{pull\_switch}, the action
of pulling a switch whose affect is unknown.  To discover what this switch does, in the form of PDDL pre- and post- conditions, we run ochre on all combinations
of ``cold" and ``red" days, pulling the switch on each one.    
The switch turns the heat on.  
So, what we should find is that only on a ``cold" day does the switch have an effect, and that effect is that a ``cold" house becomes no longer ``cold".  We start with \
simple rule ``templates", 
proposed forms of the pre- and post- conditions; the LNN machinery will set the weights to 0 for any terms that do not explain the data.

\subsection{Demonstration}
\begin{figure}[h]
\includegraphics[width=5.5in]{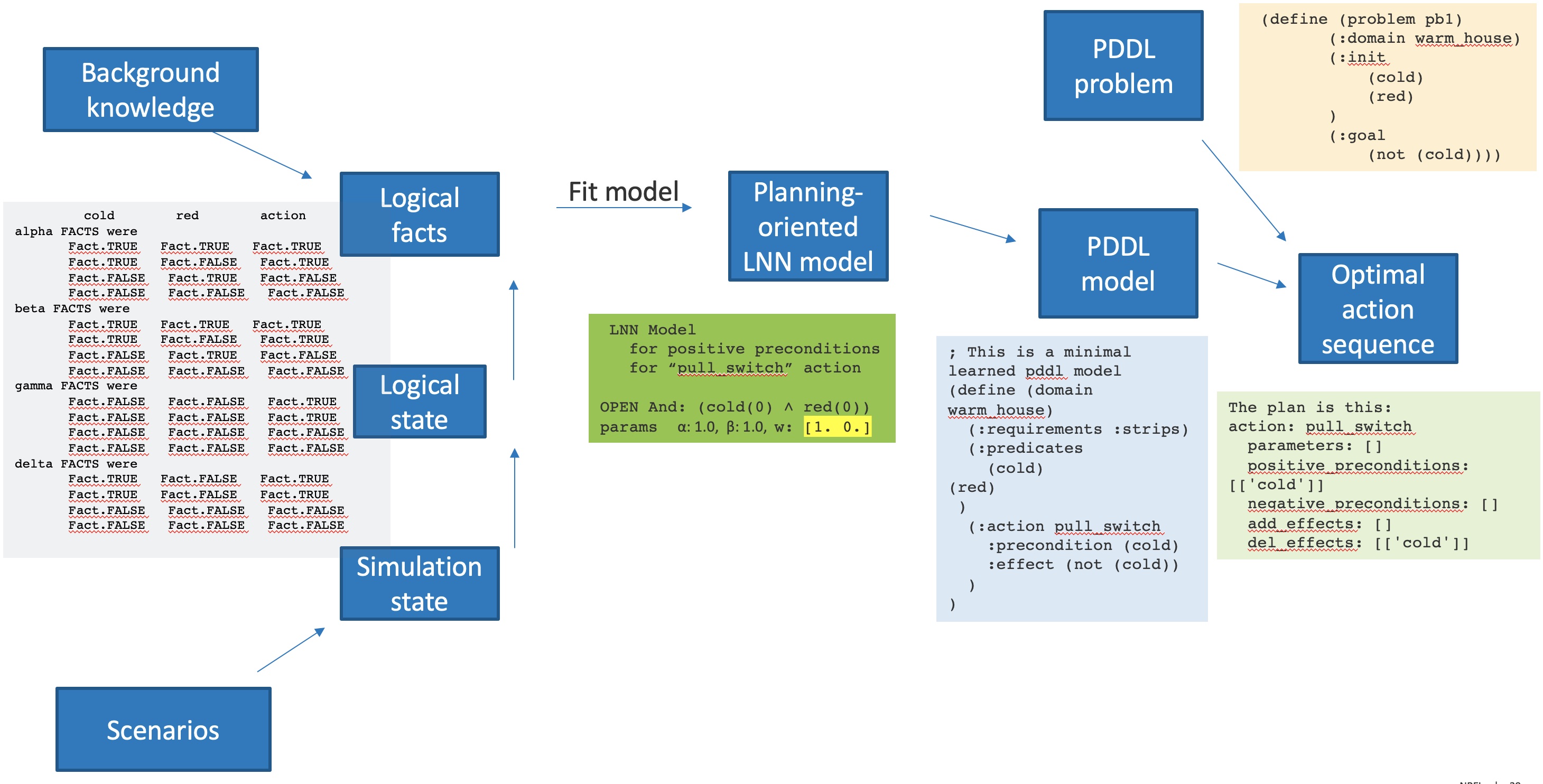}
\caption{Workflow of the OCHRE to LNN to PDDL model-based RL framework.  See text for details.}
\label{fig:loa-demo}
\end{figure}

Logically the training data is expressed by truth tables of data for each of the 4 scenarios.  For example, in Figure \ref{fig:loa-demo} (upper left), it is seen that the truth
value of ``action", i.e., whether the state changed by performing the action (pulling the switch), is just the truth value of the predicate ``cold";  ``red" is not correlated
with the truth of the conjunction.  Thus, the rule that is learned (in the green box and noting the 0 weight corresponding to ``red" in the yellow box) is simply
that the positive precondition of the \texttt{pull\_switch} action is only ``cold(x)".
Similarly, the only effect is that if a room starts as ``cold", then it becomes ``not cold".  This is expressed as the negative postcondition, seen, for example, in the full definition
of the pull\_switch action in the blue box.
Finally, together with the goal statement (in PDDL in the orange box), we submit the generated PDDL to an open-source planner, and we achieve the (quite obvious) plan 
to execute the \texttt{pull\_switch} action.

\begin{figure}[h]
\begin{verbatim}
self.vocab = {"cold": ["Temperature - Indoor (C)",  "abs", "<", mydesiredtemp], 
            "red": ["Color", "abs", "==", "red"] } }
\end{verbatim}

\begin{verbatim}
self.actions = {"turn_heat_on" : ['Load Fraction', "abs", 1, 5, 0],
                 "turn_heat_off" : ['Load Fraction', "abs", 0, 5, 1],
                 "turn_setpoint_up" : ['Setpoint', "delta", 2, 0, 0],
                 "turn_setpoint_down" : ['Setpoint', "delta", -2, 0, 0]}}
\end{verbatim}
\caption{The definitions of a vocabulary in terms of raw simulation data.  For example ``cold" is defined as a the simulation parameter ``Temperature - Indoor (C)" being less than the
python variable \texttt{mydesiredtemp}.  Similarly, the logical action ``turn\_heat\_on" is defined as setting the OCHRE control variable ``Load Fraction" to a value of 1.}
\label{fig:loa-vocab}
\end{figure}

This simple demonstration highlights several features of a LNN-based neurosymbolic planner for a simulation
tool like OCHRE.  
To summarize, the keys steps and concepts include:
\begin{itemize}
\item{define the vocabulary that allows for derivation of simulation state from logical state}
\item{define rule templates for pre- and post- conditions}
\item{probe for data across a range of scenarios (in particular, the tensor product of possible logical input values) }
\item{learn the model}
\item{generate PDDL from the model, including initial and goal states}
\item{solve the model to achieve a plan to reach the goal}
\end{itemize}

\textbf{Intepretability:} Generally speaking, though both DDTs and LNNs utilize a formulation that in principle can
result in rules and logical connectives, respectively, that are weighted combinations of various attributes and propositions,
we find the interpretability of LNNs to be superior to DDTs, simply because the learned weights almost always converge to 0 or 1.
In the case of the model-based method discussed here, we are generating data for a full (all combinations of True/False for each input
are represented in the data) and unambiguous (all truth values are strictly True or False, not in between) truth table, so the resulting
connectives do not weight the inputs but select from them those which are relevant.  The resulting rules are thus fully interpretable, as
shown in Figure \ref{fig:loa-demo}.

\section{Test Case 2: Differentiable Predictive Control with an LNN policy}
\label{sec:dpc}
\subsection{Setup}
Finally, we present an example using an LNN-based control policy in DPC, where
we implemented a simple ``HVAC" cooling problem in \pyt.  If the HVAC is on, the temperature decreases at a rate of one unit (in this example, the units are arbitrary) per time step.  If not, it increases at 0.1 unit per time step.  There is a cost imposed when the HVAC is on.
There is also a cost imposed when the temperature is above a given comfort setpoint (in our example, a value of 2). 
The goal of the exercise is to find a rule that minimizes the total cost (energy cost + discomfort cost).
We have solved this problem with LNNs in two different settings.  In the first, the cost of power is always the same.  In the second, the cost of power 
is especially high at time step 1.  In both cases, we add spurious predicates to demonstrate the ability of LNNs to learn the semantics of logical
operators from data at the same time as they learn the correct control rules. 

\subsection{Demonstration}
The LNN templates and learned rules for both examples are shown in Table \ref{tab:dpcrules}.
where ``Fake" refers to predicates that are given random
truth values during training, i.e., their values are uninformative with respect to the objective.
In both cases the LNN machinery learns to set the weights of the ``Fake" predicates to 0, resulting in the desired rules

\begin{table}
    \centering
    \begin{tabular}{|l|l|l|}
    \hline
      Template-1   & $\texttt{Implies(And(Hot(x), Fake(x)),TurnACOn(x))}$ \\
      Learned-1    & $\texttt{Implies(Hot(x),TurnACOn(x))}$  \\
          \hline
      Template-2   & $\texttt{Implies(Or(And(Hot(x), PowerCheap(x)),}$ \\
            & $\qquad \texttt{And(Fake1(x), Fake2(x))),TurnACOn(x))}$ \\
      Learned-2   &  $\texttt{Implies(And(Hot(x), PowerCheap(x)),TurnACOn(x))}$ \\
          \hline

    \end{tabular}
    \caption{Templates and learned rules for the two example DPC cases.  
    Here, `Fake' refers to a rule with random truth values that the LNN
    in both cases learns to ignore.}
    \label{tab:dpcrules}
\end{table}



\begin{figure}[h]
\includegraphics[width=3in]{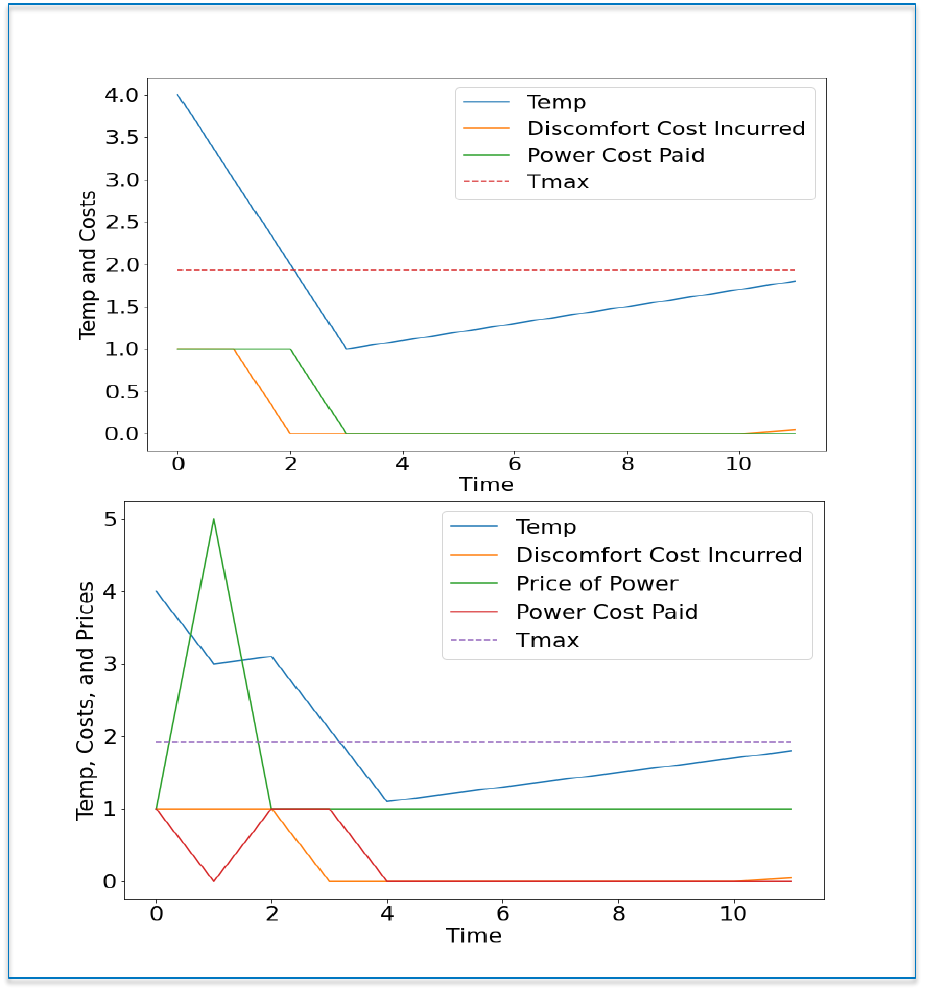}
\caption{Temperature and cost trajectories for the two test cases described in the text.  Note that in the second case 
the learned LNN rule turns the HVAC off when the power cost is high.}
\label{fig:loa-results}
\end{figure}

Figure \ref{fig:loa-results} shows the trajectories of temperature and costs for these two test cases.
In the top panel, power cost is not a variable.  We see that the rule if-cold-then-cool is correctly implemented.  The HVAC is on
through time step 2, after which the temperature has been driven low enough that it stays below the maximum desired temperature for the
rest of the episode.  This is the optimal policy, as there will always be some discomfort cost accrued due to the high initial 
temperature, but this policy drives the temperature below the threshold as fast as possible, thus incurring minimal discomfort cost.
The bottom panel adds power cost to the problem.  Here we see the rule results in a similar trajectory, with the addition of a brief
pause in the cooling while the cost of power is high.  Given the parameters of the problem (the ratio of discomfort cost to power cost), 
this is again the optimal policy.

\begin{figure}[h]
\includegraphics[width=4in]{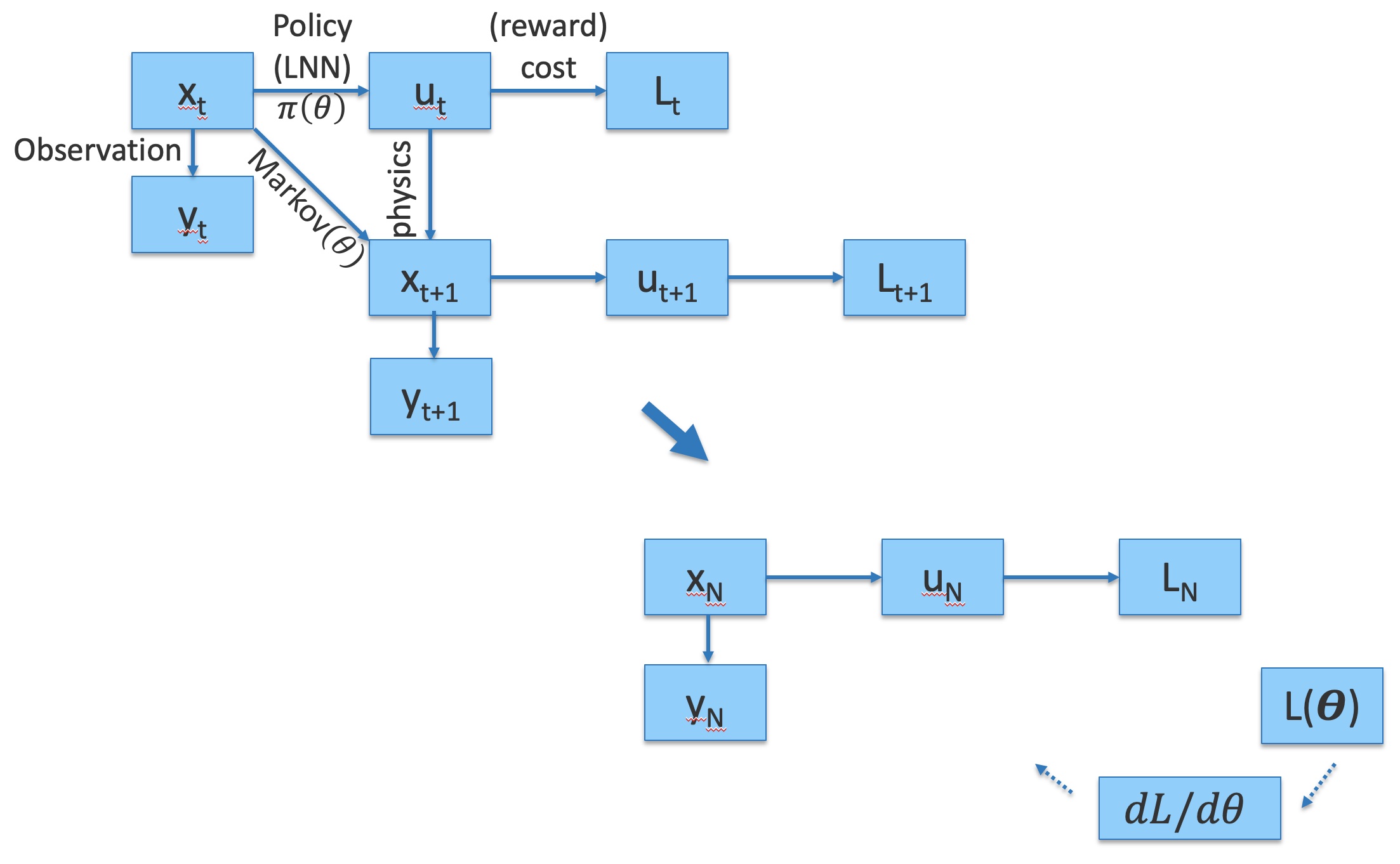}
\caption{The whole-episode computational graph resulting from integrating an LNN control scheme into a differentiable simulation.  The gradient of the episode
loss with respect to the parameters that define the LNN model is available.  Gradient descent minimizes this loss and discovers the optimal control rules.
(For demonstration purposes, the model also includes a simple hidden Markov model (HMM)).  All this machinery can be integrated and learned together via
differentiable programming.}
\label{fig:loa-graph}
\end{figure}

The issue of sharp sigmoids that was a difficulty for learning DDTs is relevant here too.  Classical logic is not differentiable.  To make it differentiable we
1) use real valued truth values, and 2) we implement formerly discontinuous if/then choices via sigmoids of a certain smoothness.
The sharper the sigmoid, the more like a true discrete decision it is.  But the sharper the sigmoid, the more numerically unstable the operations (e.g.,
gradient descent) on it become.  To overcome this problem in this case, we employ a ``smoothness" parameter during our optimization.  For the first episode,
this parameter is small (the sigmoid is very smooth), allowing very forgiving gradient descent steps.  We then increase the value of this parameter 
after each episode, so that as the parameters converge and learning occurs, we are gradually sharpening the sigmoid to the point where the
final rule represents the desired sharp if/then choice.

\textbf{Intepretability:} 
As in Section \ref{sec:mbrl}, above, we derive LNN conjunctions that have weights that are $\in \{0,1\}$, i.e.,
fully interpretable as classical logical expressions.  In this case, though, it is not as obvious that this should be the case, as we are learning
them not from fully specified deterministic truth tables, but via backpropagation through a multi-step computational graph that includes time
stepping of our simple HVAC physics.  In this proof-of-concept test case, the optimal solution is by construction interpretable, and the
differentiable-programming based method finds it.  We can imagine constructing pathological cases where the optimal connectives do not
have integer weights, but we do not pursue this issue further here.

\section{Discussion}
\subsection{Open Questions}


In this paper we have demonstrated three pathways to using differentiable interpretable models in buildings energy control, but many 
questions remain.

One of these concerns the efficacy of RBC itself. How will RBC do when the problem gets more complex?  At what point does the RBC start failing, 
or become intractable to code by hand?
This was the downfall of classical AI in the first place.  RBC was likely superior to our learned controllers only because we were
solving a problem that was too simple, i.e., in our test cases an optimal rule-based controller was readily obtained.

In more complex settings, when there is a need to generalize, we can at least imagine DRL being more adaptive;
because it is learning rich, high-dimensional representations of the observation and action spaces, it can learn
to behave subtly differently depending on inputs.  
A large neural network approximates an infinite number of conditionals (i.e.,  universal approximation).
RBC is more limited, as we would need an infinite number of rules.
This is, though, a blessing and a curse.  It constitutes a tradeoff between raw representational power and
characteristics we would like to avoid, like the need for massive data, and the lack of interpretability.
What is the best blend?  Is it a neurosymbolic method? Specifically, is it an NSAI method based on 
the sorts of differentiable interpretable representations discussed here?


The applicability of the pathways discussed here depends on scalability.   Our experience with DDTs suggests (but by no means proves)
that due to the numerical difficulties of the learning procedure, especially in the context of stochastic gradient descent in a
dynamic environment, scaling to large DDT-based rule sets may be problematic (section \ref{sec:ddtresults}).  
Conversely, while numerically the most direct, the reliance on a fully differentiable simulation of the DPC-like approach (section \ref{sec:dpc}), rarely available in practice, is also problematic.
The model-based approach of section \ref{sec:mbrl} is thus most intriguing, which presents several basic questions:
How reliably can we learn the model? How many rules can we learn? Can we learn rules with numerical parameters and more
complicated mathematical expressions that are ubiquitous in science and engineering? Can we tractably solve the resulting planning problem?
If we are only able to learn the rules incompletely, in which case a deterministic planning problem cannot be built,
we need to augment this classical planning approach with one that is more amenable to probabilistic rules.  Ingredients for 
all of these exist (e.g., the neurally-guided theorem prover, TRAIL \cite{abdelaziz2022learning}), but together are topics of further research. 

There are also many open questions regarding interpretability. 
In general, how do we resolve the tension between learnability and interpretability?
It is clear that interpretability is domain-dependent (``interpretable to whom?";
some people say NNs are interpretable; there will not be a one-size-fits-all definition).
Trying to ground our paper in BEM control makes interpretability specific.
While both of the LNN-based rule sets were manifestly interpretable (being deterministic logical expressions), the
DDTs were not as easy to interpret.  Can this be mitigated?
The discrete-to-continuous relaxation question needs more study. 
Simply adopting the ``closest" discrete solution to a continuous relaxed solution is not in general a good approach.
The analogy with integer programming is very enticing. There, in the branch-and-bound method, 
the relaxed solutions are used only as lower bounds.  A simultaneous
search for \emph{feasible} solutions results in an iterated minimization of
the gap between them and thus the discovery of the optimal feasible solution.    Is such an approach
possible (and warranted) for DDTs?
Further, we observed in our context that LNNs were more likely than DDT to result in integer weights. 
We have speculated above (sections \ref{sec:mbrl} and \ref{sec:dpc}) as to why (the optimal solution
was manifestly discrete, and we gave the system all the information to find it), but this is far from settled,
especially in the more realistic context of complex systems with noisy and incomplete data.

Finally, is there a role for large language models (LLMs) in these approaches?
Can we integrate LLMs into this process to overcome potential scalability problems of LLNs, e.g.,
to avoid starting from rule `templates' and then whittling down (via the learning process) to the correct rule set?
In this vein, we view LLMs (and neural networks in general) as `noisy idea generators'.  They
do not generate correct rules directly, necessarily, but can be used to make suggestions that are
evaluated and/or refined by our NSAI methods.
What are the advantages/disadvantages of existing knowledge representation schemas that can be integrated with LLMs for
the purpose of RBC? How do we avoid the trap of classical expert systems of needing to ask domain experts to wrack their brains for the
list of all knowledge an RBC agent might need to have?

\subsection{NSAI paradigms}
The present work exists in the context of many emerging, intertwined paradigms. 
To expand and ground the discussion we present further descriptions of a few of them.
The differentiable interpretable models that are the main subject of this paper are examples of the last of these,
namely, building models with semantic structure.

First, the notion of fast, ``System1" and slow, ``System2" thinking developed by Kahneman \cite{kahneman2011thinking} corresponds roughly to
neural versus symbolic processing.  Further reflection suggests that the operation of System2 largely consists of
repeated ``priming" of System1.  This results in a paradigm in which symbolic processing ``asks for suggestions" from neural processing,
processes these suggestions, then in an iterative fashion asks for further suggestions from the neural processing, and so on.
This is observed frequently in the current context of large language models, where humans provide the symbolic processing; 
``It's really good for spitballing ideas" \cite{npr2023}.  But we then corral these sometimes bad ideas with our own System2 thinking, 
which involves inferring
and considering meaning, reasoning about it, and making conclusions.  We repeat this process to solve problems.  There is currently
a flurry of research on how to eliminate (or reduce) the human in this neuro-symbolic (System1-System2) loop (see, e.g., \cite{jha2023neuro}).
Neurally guided search is a more prosaic version of this concept in which a neural network is learned that helps a symbolic solver
reach solutions more efficiently \cite{anthony2017thinking,ellis2018learning}.  In fact, while not always described in these terms, 
in its use of a neural network to bias the sampling 
of the classic symbolic method, Monte Carlo Tree Search (MCTS), AlphaGo Zero \cite{silver2017mastering} is another example of this paradigm.
Other works combining MCTS and neural nets include \cite{anthony2017thinking,segler2018planning}.

A second notion underlying various strands of research is the interplay between low-dimensional symbolic representations and 
high-dimensional continuous ones.  An example of a strategy that sticks to a low-dimensional representation but uses
neural nets to learn probability distributions describing the semantics of words that imply compositionality (e.g., the word `twice')
is \cite{klinger2023compositional},
which achieves perfect generalization on the SCAN \cite{lake2018generalization} and COGS \cite{kim2020cogs}
benchmarks.  Conversely, high-dimensional continuous representations are used in \cite{hersche2023neuro,trivedi2021learning}
to embed low-dimensional objects (scene descriptions and programs, respectively) in a high-dimensional space where they can be manipulated 
by linear algebra (e.g., dot products) and searched (e.g., with the cross entropy method).
An advantage of embedding is that search assuming a smooth function in a continuous domain 
can be carried out using gradient descent.  Another is that linear algebra based procedures (e.g., cosine similarity) can be used to compare  
discrete objects that are otherwise incommensurate (see, e.g., \cite{openai2023embeddings}).

Finally, examples of which constitute the main subject of this paper, 
we see the blossoming of architectures whose structure embodies rules and constraints of cognition we would like to apply. 
This line of thought connects to a longstanding debate first articulated in \cite{fodor1988connectionism} on the types of behavior neural networks are capable
of displaying, wherein Fodor and Pylyshyn argue that certain types of combinatorial behavior (in modern parlance, ``compositional generalization") can only
be achieved via systems that at some level make use of a symbol-level representation.  However, recent work by Lake 
and Baroni \cite{lake2023human} demonstrates that the combination of ``standard" transformer architecture and a clever meta-learning training regime 
achieves a rudimentary but fundamental form of compositional generalization.   Thus neural architectures converge on semantic-ness.  
This is not a new idea (e.g., Marcus clearly articulated this need in
\cite{marcus2003algebraic}). How to support symbolic processing in a learnable (thus preferably differentiable) framework is still an open question, 
but this line of attack suggests that this might
be achieved by designing neural network architectures that build in these properties, so that whatever is learned is guaranteed to have them.
  
  This idea blurs the distinction between neural and symbolic.  The final product can be seen either as a neural 
  system with symbolic semantics,
or a symbolic system implemented with neural infrastructure. 
In this light, LNNs and DDTs can be seen
as building in rudimentary forms of cognition.  In the case of LNNs, we are building in logical semantics; whatever is learned
is a series of logical statements.  For DDTs, we are building in the semantics of decision trees; whatever is learned is a form
of decision tree.

The notion of approaching semantics via complex neural network architectures is attractive (e.g., attempts to make LLMs `reason'
\cite{zhu2023large,olausson2023linc,sharan2023llm}), as it enables the critical ability to learn from data.
Behavior appearing to use semantics, however, i.e., an agent that acts ``as if" it is manipulating
symbols, is not the same as one that is \emph{actually} manipulating symbols to reach its conclusions.  
Humans manipulate symbols (albeit somewhat haphazardly).  Computers 
executing programs manipulate symbols (in a provably correct manner).  The output of transformer-based models is not the result of explicit
symbol manipulation at the level of meaningful concepts; it is the output of symbol manipulation at the level of
bits and bytes.  Arguments that
rigorous reasoning could be an ``emergent property" of a deep network beg clear demonstration of this ability. 
An
alternative path is to build the desired semantics, which implies verifiability, directly into the architectures we
learn on.  This is the tack embodied by LNNs and DDTs.  This paper is both a simple illustration and a practical roadmap of three pathways
to implement this concept in the context of reinforcement learning for building energy management.

\section{Conclusion}
The main purpose of this paper was to demonstrate three specific pathways by which a certain class of 
policy/model--namely, differentiable and interpretable, as exemplified by
LNNs and DDTs--could be utilized in reinforcement learning control of building energy systems.  These pathways were: 1) model-free RL, where we were able to
integrate DDTs directly into a standard framework, demonstrating that, because they are differentiable, such policies can be readily explored with 
off-the-shelf RL tools; 2) model-based RL, where the construction of a logic-based world model allows for directly using that model in a classical
planning context, and also allows for other existing knowledge to be directly incorporated; and 3) a method in which a fully differentiable simulation is integrated with the differentiable policy, allowing for direct gradient-based
policy optimization without the need to sample that plagues model-free RL.

\section*{Acknowledgement}
We would like to thank Alexander Gray, Asim Muniwar, Don Jovan Agravante, and Chitra Subramanian of IBM Research for their many fruitful
and illuminating discussions. 
This work was authored by the National Renewable Energy Laboratory (NREL), operated by Alliance for Sustainable Energy, LLC, for the U.S. Department of Energy (DOE) under Contract No. DE-AC36-08GO28308. This work was supported by the Laboratory Directed Research and Development (LDRD) Program at NREL. The views expressed in the article do not necessarily represent the views of the DOE or the U.S. Government. The U.S. Government retains and the publisher, by accepting the article for publication, acknowledges that the U.S. Government retains a nonexclusive, paid-up, irrevocable, worldwide license to publish or reproduce the published form of this work, or allow others to do so, for U.S. Government purposes. 
The research was performed using computational resources sponsored by the Department of Energy's Office of Energy Efficiency and Renewable Energy and located at the National Renewable Energy Laboratory.



\nocite{*}
\bibliographystyle{ios1}           
\bibliography{DINSRL}        

\end{document}